# On Planning while Learning

**Shmuel Safra**                                    SAFRA@CS.HUJI.AC.IL
*Computer Science Department*
*Hebrew University*
*Jerusalem, Israel*

**Moshe Tennenholtz**                               MOSHET@IE.TECHNION.AC.IL
*Industrial Engineering and Management*
*Technion*
*Haifa 32000, Israel*

## Abstract

This paper introduces a framework for Planning while Learning where an agent is given a goal to achieve in an environment whose behavior is only partially known to the agent.

We discuss the tractability of various plan-design processes. We show that for a large natural class of Planning while Learning systems, a plan can be presented and verified in a reasonable time. However, coming up algorithmically with a plan, even for simple classes of systems is apparently intractable.

We emphasize the role of off-line plan-design processes, and show that, in most natural cases, the verification (projection) part can be carried out in an efficient algorithmic manner.

## 1. Introduction

Suppose you find yourself in a complex labyrinth, with no recollection as to what brought you there or how to get out. You do some knowledge as to the possible outcomes of your actions (e.g., gravitation works as usual). However, several basic characteristics of your surrounding are unknown (e.g., the map of the labyrinth, or where you are in it). Your goal is to plan your way out of there while learning enough facts about your surroundings to enable that goal.

The above example is a special case of the following general setting: An *agent* $\mathcal{P}$ operating in an *environment*, is trying to achieve a given goal. At each point in time, the agent is in a specific state. The agent can be fully described by a decision procedure, which determines the next action to be taken, as a function of its history of states. An environment is taken to have some *behavior*, determining, for every state and action taken by the agent, the next state that will be reached. $\mathcal{P}$ is given a set of possible behaviors of the environment, only one of which is the actual behavior of the specific environment. We say that $\mathcal{P}$ has partial information on the behavior of the environment. $\mathcal{P}$'s goal is given as a subset of the states. Reaching any of these states is considered a success.

$\mathcal{P}$'s goal is, of course, not necessarily achievable; it may be the case that for one of the possible behaviors of the environment, there does not exist a sequence of actions that would lead $\mathcal{P}$ to a success. Moreover, even if for every possible specific behavior of the environment there exists a sequence of actions that leads $\mathcal{P}$ to its goal, it may still be the case that $\mathcal{P}$ cannot achieve its goal. For example, consider an environment with two possible behaviors $E_1$ and $E_2$. It may be the case that the only action $a$ that leads $\mathcal{P}$ to its goal when the environment follows $E_1$, leads to a state from which the goal is not achievable if the environment behaves according to $E_2$.

Nevertheless, even in the case in which $\mathcal{P}$'s knowledge of the environment is not complete, it may sometimes be possible for $\mathcal{P}$ to achieve its goal. Suppose that we change the above example so that there exists an action $c$ that, if taken by $\mathcal{P}$, leads in the case that the environment behaves according to $E_1$, to a state which is observably different from the state that results from the same action ($c$) taken when the environment behaves according to $E_2$. In addition, suppose that there exists an action $d$ that, in both cases, reverses $c$'s effects. In this case, $\mathcal{P}$ can always achieve its goal, by following the following *plan*: first take $c$, and, according to the resulting state decide whether the environment behaves according to $E_1$ or according to $E_2$. Then, take action $d$ to get back to the initial state. Finally, apply the applicable sequence of actions for either the $E_1$ or the $E_2$ case.

In general, $\mathcal{P}$ may perform some actions that reduce the number of possible behaviors of the environment (i.e., increase the knowledge that $\mathcal{P}$ has on the environment), while avoiding actions that may lead to failure in any of the still possible behaviors of the environment, according to $\mathcal{P}$'s knowledge. $\mathcal{P}$ may eventually learn enough about the behavior of the environment to choose the applicable action that leads to success. This process is referred to as *Planning while Learning*.

This paper discusses the framework of Planning while Learning, while concentrating on the *tractability* of finding a satisfactory plan (i.e., a way to achieve the goal regardless of which possible behavior of the environment is the actual one), or checking that a given plan is satisfactory. The next section defines a basic framework where Planning while Learning can be studied. In Section 3 we discuss the computational aspects we study in this paper. In particular, we distinguish between three main types of representation, and between three main computational categories. In Sections 4–5 we classify Planning while Learning based on these computational categories and representation types. In Section 6 we discuss several extensions to our basic framework. In Section 7 we put our framework and results in the perspective of related work.

## 2. The Basic Framework

Consider the following examples which have motivated our study. The first example is taken from a medical domain. Consider a trauma-care system, where there are many[1] observations that can be made on a patient's state. Actions taken by the doctor may change these observations. For example, the doctor may be able to observe whether the patient's blood pressure is high or low, and whether the patient has high or low temperature. Based on the observations made, the doctor may need to take an action, which may in turn lead to new observations. Based on these new observations, the doctor may need to choose a subsequent action, and so on. There is a list of possible injuries that the patient might suffer from, but the exact nature of the actual injury is not known apriori. Naturally, the effects of the action taken by the doctor may depend on the actual injury of the patient. The doctor needs to devise a plan that will take the patient from his initial observable state to a goal state (i.e., a "physically stable state"). The doctor can observe the patient at each point in time, and learn facts about the actual injury (i.e., the actual environment behavior) during the execution of the plan. Hence, we get a natural situation where Planning while Learning is necessary.

---

1. The term "many" will become more concrete when we discuss representation types in the following section, and will be identified with exponential in the actual representation size.

Our second example is taken from a transportation domain. Let $G = (V, E)$ be a directed graph, where the vertices denote locations in a hostile environment. The edges denote safe routes from one location to another, that are taken to be one-way (two-way routes are described as a pair of one-way routes). One-way routes in this particular domain occur as a result of the structure of the environment and of the vehicle used by the agent. Uncertainty in this transportation domain arises from the fact that there is incomplete information about the end-point of some routes originating from some particular locations (i.e., the map is partially unknown). In some cases this incomplete information concerns only a small number of locations and routes where the number of possible end-points of a route is also small.[2] An agent moving along these routes knows the possible alternatives for the structure of the environment and can identify the locations it arrives at. The objective of the agent is to reach a given target location starting from a given initial location.

The above examples are taken from real-life situations. They are typical situations of bounded uncertainty. Similar situations occur whenever we have to operate a machine that works in one of several options. In many of those cases, the possible observations can be stated, and the set of possible environment behaviors can be listed; the actual behavior, however, may be unknown apriori. Illuminating results regarding these examples are implied by our study. Nevertheless, we first have to define our basic framework.

**Definition 2.1:** An *agent-environment system* $M = (Q, \mathcal{A}, q_0, \Gamma_M)$ consists of a set of *observable states* $Q$, a set of *possible actions* $\mathcal{A}$, an *initial state* $q_0 \in Q$ and an *actual transition function* $\Gamma_M : Q \times \mathcal{A} \to Q$ that determines for each state $q \in Q$ and action $a \in \mathcal{A}$ the next state $q' = \Gamma_M(q, a)$.

Based on the above definition we can define what a Planning while Learning system is. Notice that we associate the informal term "behavior" with the term "transition function", where the actual behavior is the actual transition function.

**Definition 2.2:** A *Planning while Learning system* $S = (M, \gamma)$ consists of an agent-environment system $M = (Q, \mathcal{A}, q_0, \Gamma_M)$, and a set of *possible transition functions* $\gamma = \{E_1, ..., E_n\}$, all sharing the same set of observable states $Q$, where the actual transition function is one of these possible transition functions.

Notice that we used the term *observable states* rather than just states. An observable state of an agent is what the agent perceives at a given point (e.g., its physical location) rather than its complete state of knowledge. We assume that an agent can always distinguish between different observable states. The complete state of knowledge can be defined based on the history of actions and observable states of the agent. This history is an ordered sequence of observable states the agent visited and actions it performed. For example, if an agent performed an action $a$ that led from an observable state $s_1$ to an observable state $s_2$, and $a$ leads to $s_2$ if and only if the environment behavior is not $b$, then we can say that the agent *learned* that the environment behavior is not $b$. The agent will know that the environment does not behave according to $b$ in the state it reaches following this action, but this knowledge is not implied by the observable state $s_2$! This enables us to obtain a succinct and natural representation of agents. This type of representation has already been used in Rosenschein's situated automata (Rosenschein, 1985) and in work on reasoning about knowledge (Halpern & Moses, 1984).

---

2. In the sequel, small will be identified with polynomial in the actual representation size. In the particular application we speak about, the number of uncertain routes is logarithmic, while the number of possibilities for each such route is bounded by a small constant.

The observable states in the transportation domain example are the locations, and the actual transition function corresponds to the actual routes in the environment. In the trauma care domain, the observable states are the possible observations of the doctor, while the actual transition function corresponds to the effects of the doctor's actions given the actual injury of the patient. In both examples the agent may reach a state where it knows complex facts about the environment, based on the facts it learned by acting and observing. However, these complex states need not be represented explicitly. Further discussion of this topic can be found in the situated automata (Rosenschein, 1985; Rosenschein & Kaelbling, 1986) and knowledge in distributed systems (Halpern & Moses, 1984; Halpern, 1988) literature.

The reader should not confuse our use of automata-like structures with other common uses of it. We don't assume that an agent acts as if it is a finite-state machine, but only that the number of possible observations and possible environment behaviors it considers is finite. The agent's decisions will be based on its history of observations and actions which determine its local state (Rosenschein, 1985; Halpern & Moses, 1984) and is much more complex than its observable state. The agent's local state is not necessarily represented explicitly. This gives succinct and useful representations, as the ones discussed in Discrete Event Systems (DES) (Ramadge & Wonham, 1989) and in work in AI that incorporates uncertainty to control-theoretic models (Moses & Tennenholtz, 1991).

The above model is fundamental and some extensions of it will be discussed in Section 6. Given this model, we are now able to define the basic problem in Planning while Learning. The problem is to find a satisfactory plan that achieves a goal given any possible behavior of the environment. This problem is further discussed in the following section and investigated in subsequent sections. Similar definitions hold, and similar results can be obtained, if we require the agent to achieve its goal only in a fraction (e.g., 90%) of the possible behaviors.

**Definition 2.3:** Let $S = (M, \gamma)$ be a Planning while Learning system, where $M = (Q, \mathcal{A}, q_0, \Gamma_M)$, and $\gamma = \{E_1, ..., E_n\}$. A *goal* $g$ for the agent $\mathcal{P}$, is a subset of the states $Q$. A *plan* for an agent is a function from its history of states (in $Q$) and actions (in $\mathcal{A}$) to an action (in $\mathcal{A}$). Given a goal $g$, a *satisfactory plan* is a plan that guarantees that the agent will reach a state in $g$ starting from $q_0$, under any possible transition function in $\gamma$. A plan is called *efficient* if the number of actions that are executed in a course of it is polynomially bounded (in the representation of the Planning while Learning system).

A satisfactory plan is therefore a plan in which the agent learns enough about the environment behavior, in order to guarantee the achievement of the agent's goal. Notice that, in general, a plan might be very complex. An agent might arrive, in the course of its learning process, to situations where its goal is no longer achievable. Hence, the agent has to find a proper combination of learning and acting phases. A satisfactory plan can be viewed as a decision tree where each edge is associated with a pair of an observable state and an action to be performed in that state. This is a general representation of conditional plans. An efficient plan will therefore correspond to a decision tree of polynomial depth. Notice, however, that the size of an efficient plan may still be exponential.

## 3. A Computational Study

In the previous section we defined a basic framework of Planning while Learning, and what a satisfactory plan for an agent $\mathcal{P}$ is. In this section and in the following ones we would like to consider the complexity of finding such a plan or checking that a given plan is satisfactory. In order to discuss the issue of complexity we need to discuss our measures of complexity, and the type of representations of Planning while Learning systems we would like to look at.

### 3.1 Basic Representations

We will distinguish between three basic Planning while Learning system representations:

1. **General Representations:** Both the number of agent's observable states (i.e., $|Q|$) and the number of possible transition functions may be exponential[3] in the size of the actual representation.

2. **Quasi-Moderate Representations:** The number of agent's observable states might be exponential in the size of the system representation, but the number of possible transition functions is at most polynomial in that size. This is a most appealing type of representation for systems with bounded uncertainty (Halpern & Vardi, 1991).

   The trauma-care system mentioned in Section 2 is an example of a system with a quasi-moderate representation. In such a system we usually have a set of atomic observations (e.g., whether the blood pressure is high or low). The number of atomic observations is linear in the problem's input, but the number of possible observations (i.e., observable states which are tuples of atomic observations) is exponential. The list of possible injuries that the patient might have is usually polynomial in the problem's input. Hence, we get a quasi-moderate representation of a Planning while Learning system.

3. **Moderate Representations:** Both the number of agent's observable states and the number of possible transition functions is polynomial in the representation size. This type of representation is less general than a quasi-moderate representation, but it is still expressive and completely non-trivial as we will later discuss. The transportation domain example of the previous section is moderately represented by a graph-like structure, in cases where there are at most polynomially many alternatives for the actual structure of that graph (e.g., in a particular application, there are a constant number of possibilities to a logarithmic number of routes).

---

3. We will use the term exponential and polynomial with their standard means (i.e., polynomial and exponential in the actual representation size.)

In the remainder of this paper we use the term moderate system (resp. quasi-moderate system) to refer to a moderate representation (resp. quasi-moderate representation) of a Planning while Learning system.

## 3.2 Basic Computational Categories

Given a Planning while Learning system, there are three main computational categories that we consider.

1. **Intractable:** Checking whether a plan for the agent is satisfactory is computationally hard, and may take exponential time.

   In Planning while Learning systems that fall into this category, even the problem of representing the plan and verifying that this supplied plan is satisfactory is computationally intractable (i.e., either the space needed for representation is exponential or the verification process takes exponential time).

2. **Off-Line Tractable:** A satisfactory plan for the agent $\mathcal{P}$ has a short representation (i.e., polynomial in the representation of the problem), and checking whether a plan is satisfactory can be carried out efficiently in polynomial time.

   Systems that fall into this category may succumb to a trial and error process, in which an intelligent designer (i.e., a human) suggests some plan for solving a problem. The suggested plan is represented and verified. If it fails the verification process, then the exact failure is reported, and the designer may try to generate a new plan. This trial and error process is a typical solution for design problems. A designer is given a specific problem, and may use her experience in suggesting a plan. The plan should be represented and verified efficiently. If the plan is not satisfactory, the efficient verification process locates the failures and informs the designer, who may choose to generate a new plan, etc. This approach was at first made explicit in the AI literature, by the seminal paper of McCarthy and Hayes (McCarthy & Hayes, 1969) where it is referred to as the Missouri program. This approach is indeed the one used in many practical situations, such as the ones mentioned in the previous section. A more detailed demonstration of that idea and further discussion can be found in (Moses & Tennenholtz, 1993; Shoham & Tennenholtz, 1994).

   Hence, in systems that fall into this category, various plans can be tried in an off-line design process, supported by a computerized efficient verification procedure, which hopefully results in a satisfactory plan.

3. **On-Line Tractable:** A satisfactory plan for the agent $\mathcal{P}$ has a polynomial representation, that is not only efficiently verifiable, but can be actually computed (algorithmically) in polynomial time.

## 3.3 Basic Results

We would like to classify Planning while Learning systems based on the above categories. The following results are simple corollaries of results proved by Moses and Tennenholtz in another context (Tennenholtz & Moses, 1989; Moses & Tennenholtz, 1991; Tennenholtz, 1991) and their proof is omitted from the body of this paper.

1. Given a general representation of a Planning while Learning system, finding a satisfactory plan is PSPACE-hard. The problem remains PSPACE-hard even if we consider only efficient plans. The size of the related plan may be exponential.

2. If we do not restrict ourselves to efficient plans, then finding a satisfactory plan in quasi-moderate systems is PSPACE-hard. The size of the related plan may be exponential.

3. Finding an efficient satisfactory plan in quasi-moderate Planning while Learning systems with only one possible transition function (i.e., planning with complete information) is NP-hard. In this case, it is enough to consider plans of polynomial size.

The above results give several restrictions as to what we will be able to obtain in our study: we can not hope that finding a satisfactory plan, either efficient or inefficient, will be (even off-line) tractable given arbitrary representations of Planning while Learning systems. In addition, we can not hope that Planning while Learning in quasi-moderate representations will be on-line tractable. We remain however with several basic questions:

1. Given a quasi-moderate representation of a Planning while Learning system, is the problem of finding an efficient satisfactory plan off-line tractable?

2. Given a moderate representation of a Planning while Learning system, is the problem of finding an either efficient or arbitrary satisfactory plan tractable (either off-line or on-line)?

We will treat moderate representations first. Our results regarding quasi-moderate representations will be a simple modification of a result regarding moderate representations. We would like now to show why the problem of Planning while Learning, even in moderate representations, is non-trivial.

Consider an agent $\mathcal{P}$ who does not have complete information on the environment behavior, i.e., there may be more than one possible behavior of the environment. $\mathcal{P}$'s plan, instead of being a sequence of actions, becomes a decision tree. $\mathcal{P}$'s action, in this case, is a function, not only of the observable state, but also of the past history of $\mathcal{P}$. In the example mentioned in the introduction, where $\mathcal{P}$ has to first take the action $c$ to distinguish behavior $E_1$ from behavior $E_2$, $\mathcal{P}$'s plan has a different branch for the case the environment behaves according to $E_1$ and for the case it behaves according to $E_2$.

Note that, introducing $\mathcal{P}$'s memory as a parameter in its plan — this is essentially the difference between a sequence and a decision tree as $\mathcal{P}$'s plan — may cause an exponential blow-up in the size of that plan, and may make intractable the task of devising or verifying a plan, even in moderate representations. This holds even when we consider efficient plans! Hence, Planning while Learning even in moderate systems is completely non-trivial.

## 4. Off-Line Tractability

In this section we show that given a moderate (resp. quasi-moderate) representation of a Planning while Learning system, whenever there is a satisfactory plan (resp. an efficient satisfactory plan) for an agent, there is a satisfactory plan (resp. an efficient satisfactory plan) that can be represented in polynomial space, and can be checked in polynomial time. As we mentioned, this is a non-trivial fact even for moderate systems. We prove the result for moderate systems, and then show why it is applicable for the richer context of quasi-moderate systems.

The proof of our off-line tractability result will follow from the following lemmas.

**Lemma 4.1:** *Let $P$ be a satisfactory plan for achieving a goal $g$, in a moderate Planning while Learning system $\mathcal{S}$ with $s$ possible behaviors (i.e., transition functions), and $t$ observable states. Then, there exists a satisfactory plan $P'$ for achieving $g$ in that system, where the longest path of $P'$ is bounded by $s \cdot t$.*

**Proof:** If the agent $\mathcal{P}$ performs along any path of $P$ more than $t$ actions without learning anything (i.e., along $t$ actions the agent does not get any new information about the actual behavior), then it must visit a particular observable state twice, without getting any new information about the actual behavior, and therefore we can shrink $P$ by dropping actions which took place between these visits. We can perform this process until there will be no sequence of $t$ actions in which no learning occurs.

The learning of the agent is monotonic: whenever it learned something about the environment behavior, future information can just make this knowledge more concrete. Since the number of possible behaviors is $s$, we get that a knowledge increase can occur at most $s$ times.

Combining the above observations leads to the desired result. ∎

**Lemma 4.2:** *Let $P$ be a satisfactory plan, for a moderate Planning while Learning system $\mathcal{S}$ with $s$ possible behaviors, and where the longest path of $P$ is of length $t$. Then, there exists a representation of $P$ (in size polynomial in $s$ and $t$) such that verifying that $P$ is satisfactory can be carried out in time polynomial in $s$ and $t$.*

**Proof:** The concise representation $P'$ of $P$ consists of a table, where each entry of the table corresponds to a distinct observable history of an interaction of the agent $\mathcal{P}$ with the environment, and contains an action to be taken by $\mathcal{P}$ for that specific (partial) scenario.

The number of distinct entries in the table ($P'$) can be limited to include only the plausible distinct histories (i.e., the histories which can be generated) for the system $\mathcal{S}$ and the plan $P$. The number of such distinct histories is bounded by $s$ (the number of possible behaviors of the environment) times $t$ (the different stages in a specific interaction).

In order to verify that a plan $P'$ which is represented in that manner is satisfactory, one needs to go over all possible behaviors and for each one of them check that $P'$ leads to $\mathcal{P}$'s goal. ∎

As an immediate corollary we obtain the following:

**Theorem 4.3:** *Finding a satisfactory plan for any moderate Planning while Learning system is off-line tractable.*

Consider an agent who wishes to reach his destination in the hostile environment of Section 2. In principle, there might be exponentially-many histories of observations the agent may encounter. Nevertheless, our result says that it is enough to consider only polynomially-many of them in order to specify the appropriate plan. This is most helpful for the designer; she will be able to represent her suggested solution in a relatively concise way. If the suggested solution is not satisfactory, this fact will be efficiently detected, and perhaps can be repaired. The problem of navigation in a hostile environment we mentioned above is actually solved this way.

Notice that Lemma 4.1 is quite satisfactory for moderate systems. However, in quasi-moderate systems this Lemma is not useful, since in that case $t$ might be exponential in the actual representation size. However, the properties obtained by Lemma 4.1 can be regained by considering efficient plans. For most practical purposes, we do not lose generality by restricting our attention to efficient plans, since a planner will not be able to execute exponentially many actions in the course of a plan. Given that Lemma 4.2 does hold for quasi-moderate representations, we get:

**Theorem 4.4:** *Given a quasi-moderate Planning while Learning system, finding an efficient satisfactory plan is off-line tractable.*

This result is quite satisfactory, since quasi-moderate systems are a rich context. For example, some architectures such as the ones discussed by Brooks and his colleagues (Brooks, 1986) can be treated as quasi-moderate systems. They include a polynomial number of sensors, which correspond to an exponential number of possible observations, and are tested against a list of possible environment behaviors (i.e., the appropriate sensor-effector mechanism is checked for a list of environment behaviors). As we mentioned before, quasi-moderate systems correspond to complex systems where the number of possible worlds describing the environment is efficiently enumerable. These constitute a rich and appealing family of systems (Halpern & Vardi, 1991). Our results show, for example, that the trauma-care system discussed in the previous section can be built as an expert system that devises the next action to be performed based on the history of observations by the doctor. The problem of coming up with the plan may not be trivial, but our results show that a concise representation of a plan which is efficiently verifiable does exist whenever an efficient satisfactory plan exists. Therefore, the effort of generating the appropriate plan off-line is worthwhile.

## 5. On-Line Intractability

In this section we show that it is not likely that there is a general algorithm to come up with a satisfactory plan for any moderate Planning while Learning system, since just deciding whether such a plan exists is NP-hard. We prove the result for the basic framework of Section 2. A similar result holds regarding efficient satisfactory plans. This will imply similar results for the case of efficient satisfactory plans in quasi-moderate Planning while Learning systems, and for the extended frameworks discussed in the following section.

This result together with the results obtained in the previous section complete the classification of Planning while Learning discussed in Section 3.

**Theorem 5.1:** *Given a moderate Planning while Learning system, deciding whether there exists an (arbitrary or efficient) satisfactory plan for the agent $\mathcal{P}$ is an NP-hard problem.*

**Proof:** Given any 3-SAT formula $\varphi$, over variables $v_1, ..., v_n$ and consisting of clauses $c_1, ..., c_t$, we construct, in polynomial time, a moderate Planning while Learning system $\mathcal{S}_\varphi$, such that there exists a satisfactory plan for $\mathcal{P}$ in $\mathcal{S}_\varphi$ if and only if there exists an assignment to $v_1, ..., v_n$ that satisfies $\varphi$. Since satisfiability of a 3-SAT formula is NP-hard, this implies that deciding whether there exists a satisfactory plan, even for moderate systems, is an NP-hard problem. Our reduction will hold for the case of efficient satisfactory plans as well.

The set of observable states $Q$, in the system $\mathcal{S}_\varphi$, is $\{b, q_1, ..., q_{n+t+1}\}$. The possible behaviors of the environment are $\left\{E_{1,\bar{0}}, ..., E_{n,\bar{0}}, E_{1,\bar{1}}, ..., E_{n,\bar{1}}\right\}$ (there are $2n$ possible behaviors). The initial state is $q_1$. The set of possible actions for $\mathcal{P}$ is $\{\bar{0}, \bar{1}, a_1, ..., a_7\}$. $\mathcal{P}$'s goal is to reach the state $q_{n+t+1}$.

The state $b$ is a *black-whole*, where any action that $\mathcal{P}$ takes from $b$ results back at $b$ (which is an unsuccessful state).

From any state $q_i$, $i \in \{1, ..., n\}$, in both the cases, where $\mathcal{P}$ takes the action $\bar{1}$ and the environment behaves according to $E_{i,\bar{0}}$, and where $\mathcal{P}$ takes the action $\bar{0}$ and the environment follows the behavior $E_{i,\bar{1}}$, the resulting state is $q_{n+t+1}$ ($\mathcal{P}$'s goal). For all other behaviors, if $\mathcal{P}$ takes the actions $\bar{0}$ or $\bar{1}$ from state $q_i$, the resulting state is $q_{i+1}$. (Taking the actions $a_1, ..., a_7$ leads to the state $b$).

For any clause $c_j$, with each assignment (to the variables mentioned in $c_j$) that satisfies $c_j$, we associate one of the actions $a_1, ..., a_7$ (a clause with 3 variables has 7 satisfying assignments to its variables). If the observable state is $q_{n+j}$, and $\mathcal{P}$ takes the action $a_k$, which is associated with an assignment that assigns $0$ $(1)$ to variable $v_l$, and the environment behaves according to $E_{l,\bar{1}}$ $(E_{l,\bar{0}})$, the resulting state is $b$ (hence $\mathcal{P}$'s goal is not achievable anymore); taking the action $a_k$ from the state $q_{n+j}$, under other possible behaviors, leads to state $q_{n+j+1}$.

We show now that if $\varphi$ is satisfiable then there exists a satisfactory plan for $\mathcal{P}$ in $\mathcal{S}_\varphi$. Let $S: \{1, ..., n\} \to \{0, 1\}$ be an assignment to variables $v_1, ..., v_n$, that satisfies $\varphi$. We construct a plan $P_S$ for $\mathcal{P}$ as follows: in the $i^{\text{th}}$ step, the agent takes the action $\bar{0}$ or $\bar{1}$ depending on the value of $S(i)$. Then, in step $n + j$, the agent takes action $a_k$, that corresponds to the restriction of $S$ to the variables that appear in $c_j$. It is easy to see that $P_S$ leads to success regardless of the actual environment behavior.

On the other hand, given a satisfactory plan $P$ for $\mathcal{P}$, we show there exists an assignment $S_P$ that satisfies $\varphi$. $S_P$ is constructed according to the first $n$ steps of $P$ (for the behaviors that did not reach success yet) — which must be either $\bar{0}$ or $\bar{1}$. $S_P$ satisfies $\varphi$, otherwise there would be a clause $c_j$, such that any assignment that satisfies $c_j$, contradicts the assignment of $S_P$'s value to one of the variables $v_l$, which would cause failure, on the $(n+j)^{\text{th}}$ step, for either behavior $E_{l,\bar{0}}$ or $E_{l,\bar{1}}$. ∎

# 6. Extending the Framework

The previous sections introduced and investigated a general framework of Planning while Learning. A major feature of the model discussed in the previous sections is that the agent does not affect the environment behavior. This is quite natural in many applications. In many cases we may wish to consider a particular set of possible worlds (i.e., behaviors, transition functions), and there is no reason to assume they may change, given that a possible world specifies a full transition function. An interesting extension results from relaxing this feature. For example, in the transportation domain described in Section 2, one may wish to consider a case where moving along a particular route prevents future movements along other routes. This is due to the fact that movements along some routes may reveal the agent's existence to an enemy and will prevent the agent's movement along some routes that are under the enemy's control. Another interesting extension we would like to consider is the case of a multi-agent system instead of a single-agent one.

Both of the above extensions are strict generalizations of our basic framework. Therefore, our on-line intractability results hold in the extended frameworks as well. However, questions regarding off-line tractability should be carefully considered. We will define these extended frameworks and investigate the off-line tractability of the related problems.

## 6.1 Dynamic Behaviors

**Definition 6.1:** An *extended Planning while Learning system* $S_e = (Q, \mathcal{A}, q_0, B, b_0, \Gamma_e)$ consists of a set of *observable states* $Q$, a set of *possible actions* $\mathcal{A}$, an *initial agent's state* $q_0 \in Q$, a set of *environment behaviors* $B$, an *initial environment behavior* $b_0 \in B$, and a *global transition function* $\Gamma_e: Q \times B \times \mathcal{A} \to Q \times B$, that determines for each state $q \in Q$, behavior $b \in B$, and action $a \in \mathcal{A}$, the next state and behavior $(q', b') = \Gamma_e(q, b, a)$.

Notice that in extended Planning while Learning systems, the global transition function may change the behavior (i.e., the actual transition function) of the environment.

The definition of a goal and of a satisfactory plan will remain as in the basic framework. More specifically, we assume that the agent does not initially know the identity of $b_0$, but wishes to devise a plan that will succeed regardless of the identity of $b_0$. The agent however knows $\Gamma_e$. These assumptions will capture Planning while Learning in the extended framework. A moderate (resp. quasi-moderate) extended Planning while Learning system is a Planning while Learning system in which the number of elements in $B$ is polynomial, and the number of elements in $Q$ is polynomial (resp. exponential) in the size of the actual representation. The meaning of these definitions is as in the basic Planning while Learning framework.

Unfortunately, Lemma 4.1 does not hold even for moderate extended Planning while Learning systems. However, as we mentioned in Section 4, the properties obtained by Lemma 4.1 can be regained by considering efficient plans. For most practical purposes, we do not lose generality by restricting our attention to efficient plans, since a planner will not be able to execute exponentially many actions in the course of a plan. As we mentioned before, blow-up in the size of satisfactory plans may still be possible, even if we restrict ourselves to efficient plans only. We make no assumptions about the size of the related decision tree.

Fortunately, Lemma 4.2 does hold for extended Planning while Learning systems. The proof of this lemma for the extended framework is similar to its proof in the basic framework. Combining the above we get:

**Theorem 6.1:** *Given a quasi-moderate extended Planning while Learning system, finding an efficient satisfactory plan is off-line tractable.*

## 6.2 Multi-Agent Systems

Another interesting extension is concerned with the case where there is more than one agent in the system. For ease of exposition, we will assume that there are two agents that generate actions.[4]

An interesting feature of the multi-agent case is that an agent might not be familiar with the goal and the initial state of the other agent. Hence, Planning while Learning refers now to the case in which an agent tries to achieve its goal while learning about the behavior of the environment, and about the goals and initial states of other agents.

**Definition 6.2:** A *multi-agent Planning while Learning system* is a tuple
$S_m = (Q_1, Q_2, \mathcal{A}, q_0^1, q_0^2, B, b_0, \Gamma_m)$ where $Q_i$ is a set of *observable states for agent i*, $\mathcal{A}$ is a set of *possible actions*, $q_0^i \in Q_i$ is the *initial state of agent i*, $B$ is a set of *environment behaviors*, $b_0 \in B$ is an *initial environment behavior*, and $\Gamma_m : Q_1 \times Q_2 \times B \times \mathcal{A}^2 \rightarrow Q_1 \times Q_2 \times B$ is a *global transition function* that determines for each pair of states $q_1 \in Q_1, q_2 \in Q_2$, behavior $b \in B$, and a joint action of the agents $(a_1, a_2) \in \mathcal{A}^2$, the next observable states of the agents and the next environment behavior: $(q_1', q_2', b') = \Gamma_m(q_1, q_2, b, a_1, a_2)$.

Each agent has its own goal, and its plan is a decision tree that refers only to that agent's observable states. The definitions of moderate and quasi-moderate representations are straightforward generalizations of their definitions for extended Planning while Learning systems. In addition, we assume that each agent can start in one of polynomially many initial observable states, and may have one of polynomially many goals it might be required to achieve. Nevertheless, each agent may not know what the exact initial state of the other agent is, and what the exact goal of the other agent is. We are interested in satisfactory *multi-agent plans*. Formally, we have:

**Definition 6.3:** Given a multi-agent Planning while Learning system, a *multi-agent plan* is a pair of sets of plans, one set for each agent. Let $Goal_i$ denote the set of plans for agent $i$. A multi-agent plan is *satisfactory* if for each agent $i$ and for each possible goal $g$ of agent $i$, there is a plan in $Goal_i$, that achieves $g$ starting from any possible initial state, regardless of the plan (in the corresponding $Goal_j$) and initial state of the other agent, and regardless of the initial behavior of the environment. An *efficient satisfactory multi-agent plan* is a satisfactory multi-agent plan that consists of plans which are decision trees of polynomial depth.

The above definition captures intuitive situations of Planning while Learning in multi-agent domains. Assume for example that there are two forces that have to move in the hostile environment of Section 2. They start moving on 5AM, and need to reach their destinations by 9PM. Nevertheless, they can not be sure about the exact initial location of each other and about each other's destination. What the commander attempts to do in that case, is to devise a master-plan that should be good for all goals, initial locations, and environment behaviors. This master-plan is the satisfactory multi-agent plan we look for. Notice that movements of one agent may affect the behavior of the system and the results of other agents' movements. It is easy to see that similar scenarios occur in the trauma-care example and in many other natural systems.

---

4. Our discussion and results hold for any constant number of agents.

We now show that our off-line tractability result can be extended to the multi-agent case as well. We will use the following two lemmas.

**Lemma 6.2:** *Given a quasi-moderate multi-agent Planning while Learning system, where each agent has only one goal, if an efficient satisfactory multi-agent plan for achieving these goals exists, then there exists such an efficient satisfactory multi-agent plan that can be encoded in polynomial space, and be verified in polynomial time.*

**Proof:** In this case each agent knows the goal of the other agent, and hence it is clear that it might learn only facts about the possible initial states and behaviors.

Given that there is only a polynomial number of possible initial states and environment behaviors, and given the polynomial bound on the depth of the plans, there are only polynomially many sequences of observations (each of which of polynomial length) of each agent that are of interest (as in Lemma 4.2). Hence, we can encode, in polynomial space, a decision table for each agent mentioning only these sequences, and check, in polynomial time, whether it determines a satisfactory multi-agent plan. ■

**Lemma 6.3:** *Given a quasi-moderate multi-agent Planning while Learning system $S$, where each agent has $n$ possible goals (where $n$ is polynomially bounded in the actual representation size), there exists a quasi-moderate multi-agent Planning while Learning system $S'$ (where quasi-moderate refers to the actual representation size of the original system $S$), with a unique goal for each agent, such that there exists an (efficient) satisfactory multi-agent plan in $S'$ if and only if there exists an (efficient) satisfactory multi-agent plan in $S$.*

**Proof:** $S'$ will be built as follows. The observable states of agent $i$ in $S'$ will be the cartesian product of the observable states of agent $i$ in $S$ with the set of states:
$\{start_i, observe_{i_1}, \ldots, observe_{i_n}, goal_i\}$. The initial state of agent $i$ in $S'$ will be taken to be the pair consisting of its initial state in $S$ and $start_i$, and its goal is taken to be the set of states in which $goal_i$ is a component. The environment in $S'$ will be a cartesian product of the behaviors in $B$ with two sets $G_1$ and $G_2$, where $G_i$ has $n$ distinct elements: $\{g'_{i_1}, \ldots, g'_{i_n}\}$.

Agent $i$ will have a distinguished action, called $observe - goal_i$, which he must execute in its initial state. The state transition function will be as in $S$, but when $i$ performs $observe - goal_i$ its "new component" in the cartesian product (and only it) will change; the change will be to $observe_{i_j}$ if and only if the projection of the initial behavior on $G_i$ is $g'_{i_j}$. In addition, assume that $\{g_{i_1}, \ldots, g_{i_n}\}$ are the possible goals for agent $i$ in $S$, then the transition function in $S'$ will change the new component of the observable state to $goal_i$ if and only if the new component of the environment is in state $g'_{i_j}$, and a state satisfying $g_{i_j}$ has been reached.

The above transformation from $S$ to $S'$ makes the identity of an agent's goal a component of the initially unknown behavior. However, agent $i$ and no other agent will observe its goal after its first action. It is easy to see that the above transformation keeps the system quasi-moderate, and that there exists a satisfactory multi-agent plan in $S$ if and only if there exists such a plan in $S'$, where in $S'$ each agent has only one possible goal. ■

Combining the above lemmas we get:

**Theorem 6.4:** *Given a quasi-moderate multi-agent Planning while Learning system, finding an efficient multi-agent satisfactory plan is off-line tractable.*

The above proof shows that Planning while Learning is off-line tractable in multi-agent cases such as the ones described above. Given the structure of the above lemmas, it is easy to prove similar results for other contexts where there is a polynomially bounded uncertainty about a multi-agent system. For example, if we would like to find a multi-agent plan where crash failures of agents might occur (in that case the faulty agent might not achieve its goal, but we require that the other agent will still be able to achieve its goal), then we can show that this problem is off-line tractable, using the above techniques.

## 7. Related Work

Early work in the area of planning was devoted to various cases of planning with complete information (see (Allen, Hendler, & Tate, 1990) for many papers on that topic). As research in this area progressed in various directions, several independent works observed that the assumption that a planner has complete information is unrealistic for many situations; the sub-area that treats that aspect of planning is usually referred to as planning in uncertain territories.

Examples of research in this sub-area include work concerning knowledge and action (Moore, 1980; Halpern, 1988), work on conditional and reactive plans (Dean & Wellman, 1991) and work on interleaving planning and execution (Ambros-Ingerson & Steel, 1988). The reactive approach is proposed as a tool in the control of robots operating in uncertain environments, and in the design of real-life control architectures that would be able to react in a satisfactory manner, given unpredicted events (Brooks, 1986). The interleaving of planning and execution may sometimes be a useful alternative to conditional planning. However, in many realistic domains there is a need to consider a whole or large portion of a plan before deciding on an action. This is the case in the transportation domain and the trauma-care domain we discussed. Nevertheless, we see the interleaving of execution with Planning while Learning a promising direction for future research.

Research in the direction of conditional plans deals with plans in which the outcome of the agent's action may affect the next action taken by the agent. Theoretical work on this issue is mainly devoted to aspects of reasoning about knowledge and action (Moore, 1980; Halpern, 1988; Morgenstern, 1987), and to the logical formulation of conditional plans (Rosenschein, 1981). Specific mechanisms to construct conditional plans in which observable events and tests are explicitly declared are discussed as well (Wellman, 1990). These as well as the more classical work on conditional plans (Warren, 1976), and work that followed and extended it in various directions (Peot & Smith, 1992; Etzioni, Hanks, Weld, Draper, Lesh, & Williamson, 1992) have not concentrated on general computational aspects of Planning while Learning. Our work does not concentrate on specific mechanisms for the construction of conditional plans; Rather, it concentrates on general computational aspects of conditional planning. Some recent work has also been concerned with computational aspects of conditional plans, but concentrated on several natural pruning rules that can be used in the construction of conditional plans (Genesereth & Nourbakhsh, 1993).

Our approach crystallizes the notion of an agent interacting with an environment, and having to come up with a conditional plan, which would lead to the agent's goal in every possible behavior of the environment (possible world). We are mainly concerned with general computational aspects of Planning while Learning, and classify Planning while Learning based on several computational categories and representation types.

Another suggested approach for planning in uncertain environments, whose applicability aroused quite a heated discussion recently, is referred to as universal plans (Schoppers, 1987). A universal plan is one in which the reaction of the agent to every possible event of the environment is specified explicitly. Our results isolate general classes of systems in which the agent's actions can be specified explicitly in an efficient manner, in order to enable automatic verification. Furthermore, systems that do not fall into the above classes may be intractable even if the agent has complete information on the environment behavior.

Other somewhat related work is concerned with planning routes where the geography is unknown (Papadimitriou & Yannakakis, 1989; Mcdermott & Davis, 1984). For example, one may be interested in finding a route leading from one city to another without access to an appropriate map. This work may be viewed as a special case of the general framework of Planning while Learning. Work on the design of physical part orienters (belts, panhandlers) that accept an object in one of several possible orientations and output it in a predetermined orientation (Natarajan, 1986) may also be viewed as a special case of our framework.

Our work is concerned with the off-line and on-line tractability of Planning while Learning. This relates it to work concerned with the tractability of different types of planning (Erol, Nau, & Subrahmanian, 1992; Bylander, 1992). This work mainly concentrated on on-line tractability of a single-agent planning with complete information. Our work concentrates on general computational aspects of planning with incomplete information, considers also multi-agent situations, and discusses both on-line and off-line tractability. Recall that off-line design and tractability, although considered an attractive option (McCarthy & Hayes, 1969), has been almost neglected in the recent years (but see (Moses & Tennenholtz, 1993; Shoham & Tennenholtz, 1994)).

Research on inference of finite-automata (Rivest & Schapire, 1987, 1989) assumes an agent that tries to infer the structure of an automaton. The agent is given a limited access to the automaton, and is expected to gain enough information to deduce the complete structure of the automaton. By contrast, in the framework discussed in this paper, the agent needs only gain information that would help in reaching the given goal. Therefore, in what is probably a most natural case, the automaton is fairly complicated, thus learning its complete structure is computationally infeasible. However, being only interested in a specific goal, one may be able to obtain the necessary information, and succeed in that goal. In addition, work on computational learning assumes that the given automaton is fully connected, to enable reaching any state of the automaton and eliminating the need of avoiding states from which other states are not reachable. This assumption — that the automaton is fully connected — may very well be false in many real-life applications.

The part of our work which discusses multi-agent plans is related to issues in distributed AI (Bond & Gasser, 1988) and to the complexity of multi-agent planning (Tennenholtz & Moses, 1989); we investigate the computational difficulty that arises due to uncertainty concerning the activities of an additional agent(s).

As far as related representations are concerned, the model we present is different from classical representations in the spirit of STRIPS. It is a classical Discrete Event Systems model (Ramadge & Wonham, 1989). The general connection between planning and control theory has been discussed in previous work (Dean & Wellman, 1991). In addition, Tennenholtz and Moses show a reduction from control-theoretic models as the ones we discuss to the more classical STRIPS-like representations (Tennenholtz & Moses, 1989). They show how a typical STRIPS-like representation can be reduced to a quasi-moderate representation. However, the control-theoretic representations we considered are conceptually different from classical planning models, due to the fact that they model explicitly the possible observations of agents and the effects of actions given different environment behaviors, rather than represent general facts about an environment. The *local* (or *mental*) state of an agent, which is the general agent's state discussed in the AI literature (Shoham, 1990), will not be represented explicitly in our representation and will be built implicitly based on the agent's actions and observations. Hence, the most appropriate similar model of knowledge representation in AI is the situated automata (Rosenschein, 1985). Notice that, in general, the number of local states an agent might reach is exponential in the number of its observable states.

## 8. Conclusions

A useful planning system needs to have three essential properties. First, it should supply a mechanism for the generation of plans. Second, it should supply a concise way for representing plans. Third, it should supply an efficient mechanism for the verification of plans or for testing candidate plans.

In this paper we concentrate on planning in uncertain territory, where the agent has only partial information on the environment behavior. We show that it is intractable to build a useful planning system even for moderate representations (i.e., representations in which the number of observable states and possible behaviors is polynomial in the actual representation size). However, our positive results show that it is possible, in moderate and quasi-moderate representations (where the number of observable states might be exponential), to satisfy the 2nd and 3rd properties mentioned above. Hence, off-line design becomes tractable, as discussed and demonstrated in the paper.

Notice that if we consider quasi-moderate systems and efficient plans, which is a most natural situation, our results imply that Planning while Learning is as efficient as planning with complete information. Both are off-line tractable and on-line intractable. However, in moderate systems, planning with complete information is quite trivial (this is the case of graph search (Aho, Hopcroft, & Ullman, 1974)), while in that case we show that Planning while Learning is NP-hard. More generally, we obtain a complete classification of Planning while Learning systems based on several representation types and computational categories. In addition, we discuss extensions of Planning while Learning, such as Planning while Learning in multi-agent domains.

The framework of Planning while Learning is a general framework where planning in uncertain territory can be studied. The introduction of this framework, and the related (positive and negative) results, facilitate that study.

## Acknowledgements

We would like to thank Dan Weld, and three anonymous reviewers, for their helpful comments.